# Implementation of WSN which can simultaneously monitor Temperature conditions and control robot for positional accuracy

Sharul Agrawal, Mr. Ravi Prakash, Prof. Zunnun Narmawala
Computer Science & Engineering, Institute of Technology, Nirma University, Ahmedabad, India 382 481
Engineer SF, Institute for Plasma Research, Gandhinagar, India 382 428
Computer Science & Engineering, Institute of Technology, Nirma University, Ahmedabad, India 382 481

*Abstract—* Sensor networks and robots are both quickly evolving fields, the union of two fields seems inherently symbiotic. Collecting data from stationary sensors can be time consuming task and thus can be automated by adding wireless communication capabilities to the sensors. This proposed project takes advantage of wireless sensor networks in remote handling environment which can send signals over far distances by using a mesh topology, transfers the data wirelessly and also consumes low power. In this paper a testbed is created for wireless sensor network using custom build sensor nodes for temperature monitoring in labs and to control a robot moving in another lab. The two temperature sensor nodes used here consists of a Arduino microcontroller and XBee wireless communication module based on IEEE 802.15.4 standard while the robot has inherent FPGA board as a processing unit with xbee module connected via Rs-2332 cable for serial communication between zigbee device and FPGA. A simple custom packet is designed so that uniformity is maintained while collection of data from temperature nodes and a moving robot and passing to a remote terminal. The coordinator Zigbee is connected to remote terminal (PC) through its USB port where Graphical user interface (GUI) can be run to monitor Temperature readings and position of Robot dynamically and save those readings in database.

## I  INTRODUCTION

Wireless Sensor Network (WSN) are a trend of the last few years due to the advances made in the wireless communication, information technologies and electronics field [1].Wireless sensor networks is particularly useful for areas which are difficult to hardwire , where there is no easy access to a tethered power source and where human intervention is not possible. It can allow for easy deployment by vastly reducing the amount of cable that must be run. Sensor nodes or *motes* in WSNs are small sized and are capable of sensing, gathering and processing data while communicating. .The brain of each WSN node is the microcontroller which processes readings from its own sensors. Also a central base station is needed from where full operation and monitoring can be carried out.

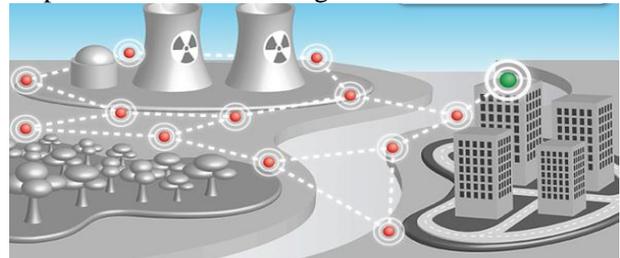

Fig 1: Scenario for WSN

In this prototype system, we develop an embedded wireless sensor prototype system for temperature monitoring in labs of Institute for plasma Research(IPR). We used Xbee modules based on IEEE 802.15.4 standard and Arduino Board which comes with ATmega168 or 328 for easy interfacing with the XBee module and for easy programming (in C) of the microcontroller. The Arduino boards come with a library for interfacing with XBee module and for dealing with analog or digital inputs and outputs.

The remaining of the paper is organized as follows.
In Section II details about zigbee Ad-hoc network and various logical devices are described . Section III describes the implementation part. Section IV describe the Application of system and Section V  conclude the paper with results.

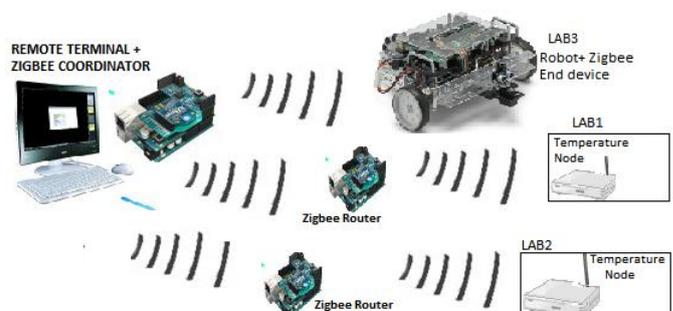

Fig2: Overview of system

## II CREATION OF ZIGBEE AD-HOC NETWORK

ZigBee is a wireless technology used for implementing WPAN. It is based on the IEEE 802.15.4 standard which defines the PHY (Physical) and MAC (Media Access Control) layers of the ZigBee protocol stack. The XBee module used here is shown in Fig 3 manufactured by Digi and X-CTU software is used for configuring XBee.

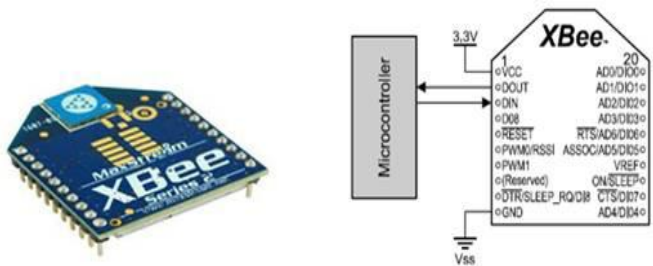

Fig 3: xbee module and pin diagram[2]

The IEEE 802.15.4 standard defines 3 classes of logical devices- [3]

1) **Zigbee Coordinator(ZC)**
   - This device starts and controls the network.
   - Selects a channel and PAN ID
   - Allow routers and end devices to join the network
   - Buffers data packets if End devices are in sleeping mode
2) **Zigbee Router(ZR)**
   - Extend network area coverage
   - dynamically route around obstacles, and provide backup
   - routes in case of network congestion or device failure.

3) **Zigbee End Device(ZED)**
   - communicate with ZC or ZR but cannot transmit information to other devices
   - Asleep most of the time

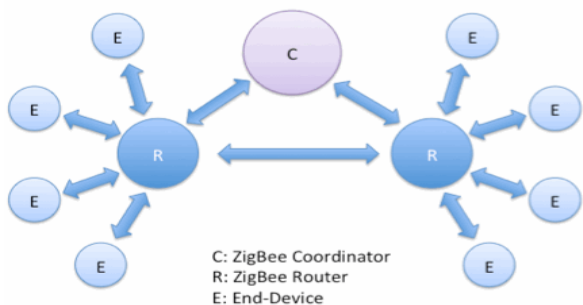

Fig4: Zigbee Networks [2]

For any Zigbee device to join a network, they need to be programmed to include its own 64 bit personal Area network ID(PAN ID). This is done through the X-CTU software with AT command. The XBee firmware includes a number of AT commands that can be used to configure the XBee joining parameters (i.e. scan channels, PAN ID, permit join, security settings, 64 bit Destination address DH and DL respectively etc) using X-CTU software..

## III IMPLEMENTATION

### A. OVERVIEW

- For the given application, a Personal Area Network is created with the help of 2 temperature sensor nodes, one robot, two routers and one co-ordinator as shown in fig 2.
- Two sensor nodes are kept in two different labs which are used for temperature sensing and in another lab LabVIEW FPGA Starter kit robot from National Instrument(NI) [6] having inherent position sensor(Encoder) is controlled and monitored dynamically.
- The remote nodes having Zigbee modules will also be put into sleep mode for synchronization among nodes.
- The microcontroller code for the two remote nodes is written in Arduino software[5] and FPGA code to determine the position of robot is written in LabVIEW 2011 software using serial programming. The application is then deployed on sbRIO board mounted on Robot as shown in fig 5.

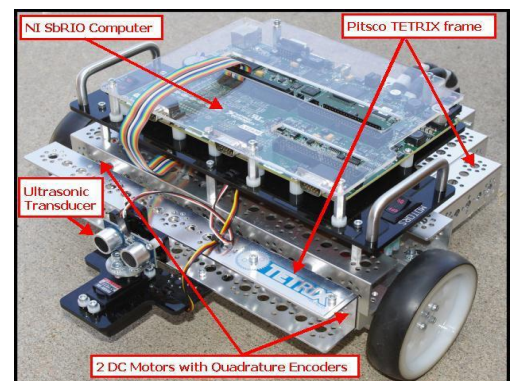

Fig5: NI Starter kit 2.0[4]

- At last development of a Graphical user interface which can be run on remote terminal to monitor Temperature readings and position of Robot dynamically and save those readings in database.

B. HARDWARE DESIGN

1. TEMPERATURE NODE

A sensor node generally consist of the following: controller, memory, sensors , communication module and a power supply . Keeping in mind the need of our application we designed our own custom nodes comprising following components in it.

**Controller -** Atmega328(Arduino) microcontroller.
**Software-** Arduino IDE
**Sensors -**LM35 temperature sensor.
**Communication-** Zigbee S2 module on each node.
**Power Supply-** 9V Batteries on each node.
**Power saving -**Sleep modes for microcontroller and Zigbee when not transmitting data.

The following figure shows the design of temperature node.

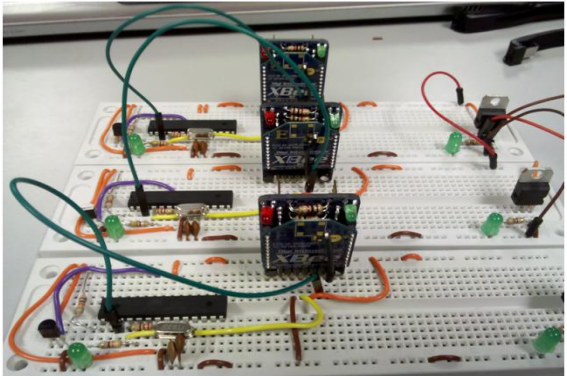

Fig6: Custom sensor node

One sensor node is kept in Electrical lab for monitoring temperature and the other is kept in Cryogenic lab as shown in figure7 below.

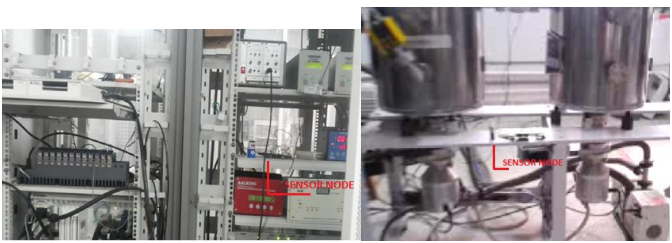

Fig 7: Node placed in Electrical and Cryogenic Lab

C. CUSTOM PACKET

In order to reliably send data between the coordinator and nodes a custom packet was created as shown in fig8. The first byte is the header byte with a fixed value of 0xFF i.e 255, the next is the node ID, then it is followed by the data bytes, and the last byte is a checksum byte. 2 Nodes are designed each with an ID 1 and 2 respectively. Once the program sees a header byte '0xFF', it knows the beginning of packet and it does the appropriate calculations to determine the ID and temperature of the remote nodes. It also does the checksum calculations to check if there any errors in the transmission adding ID and Temp value .

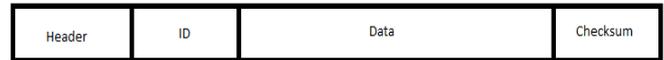

Fig8: Data packet

Temperature Sensor (LM335A) is a semiconductor [8] which measures temperature and displays the information in the voltage form. The output from the temperature sensor is analog but is then sampled and quantized (A/D converted) by the Arduino. Hence, we can realize temperature in degree Celsius by calibrating it by using the equation:

Temperature = 5 *(analogRead *100 )/1024 ) [8]

The Atmega328 microcontroller is programmed using Arduino IDE software that utilizes C language. The screenshot of Arduino microcontroller and software can be seen in fig below.

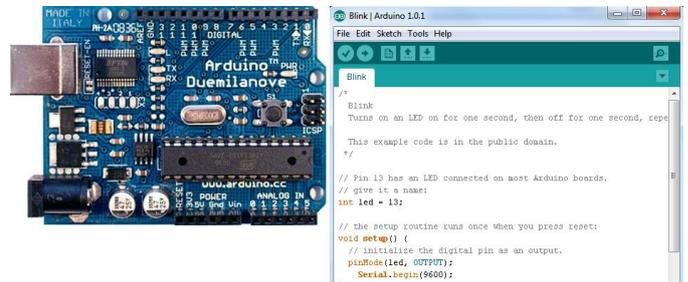

Fig9: Arduino Duemilanove and software[5]

The flowchart shown in fig10 justifies the steps followed by the two remote nodes.

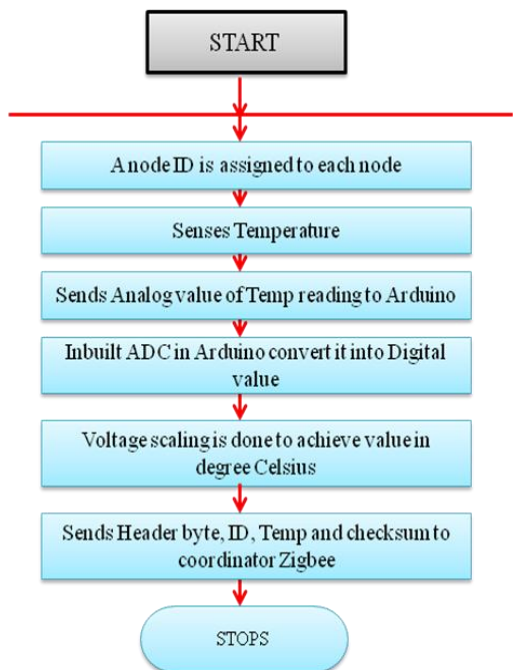

Fig10: Remote Node Process

This temperature reading along with Header byte, ID and checksum byte is then send by End device Zigbee to coordinator zigbee which performs the steps as depicted by flowchart in fig11.

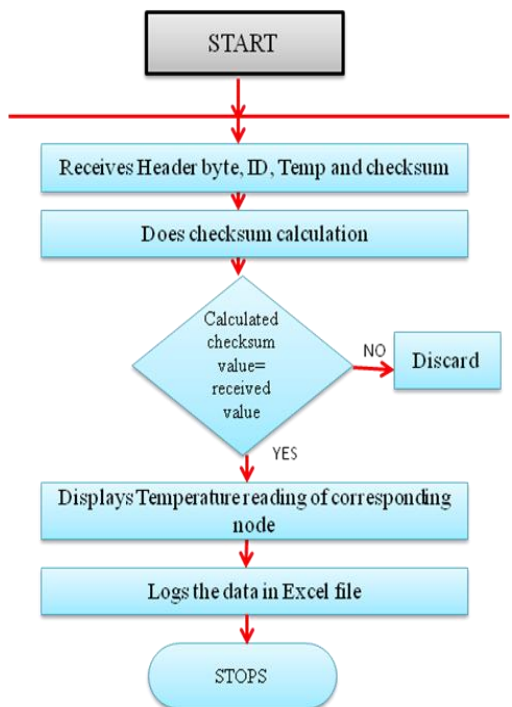

Fig11: Coordinator Node Process

### D. POSITION OF ROBOT

On the other side of network a LabVIEW FPGA Starter kit Robot 2.0 manufactured and designed by National Instruments (NI) [6] is tested to move to target x-y position given by user wirelessly and to monitor its position remotely. For example when a packet containing data of form (2,1) is send from coordinator Xbee, the end device Xbee connected to serial port of sbRIO board of robot senses it and pass it to the serial port of board through which it is connected as shown in fig 12.The serial programming in LabVIEW is done in such a way that it keeps on polling the port and as soon as it receives any valid data, the robot starts moving in respective X direction ie. uptil 2m is reached, moves 90 degree and then moves in Y direction till 1m is reached.

**Controller –** FPGA
**Software-** LabVIEW 2011
**Sensors -** Encoder as position sensor
**Communication-** Zigbee S2 module connected with robot via RS-232
**Power Supply-** Inherent in Robot.

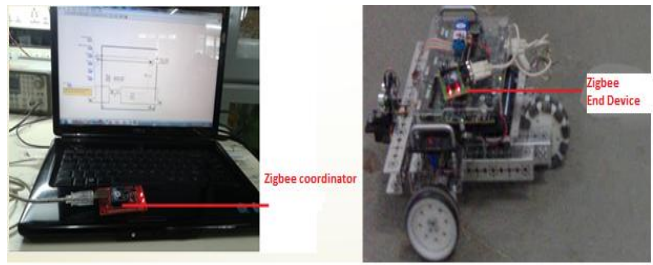

Fig12 system oveview

For estimating position of robot odometric method of localization (Dead Rocking ) [4] is used in this experiment. It calculates the position of robot using angular and forward velocity.

1) **Takes Feedback of Robot Velocity:** Encoder is a Electromechanical device which coverts mechanical rotation of shaft into the digital pulses. 2 Quadrature encoder attached with the DC motor of robot as clear from fig 5 is used for taking velocity feedback. By Counting no. of pulses , total no. of rotation of shaft is known and from that we can get both forward and angular velocity. Final position is estimated using this velocity [4]. The program was tested upon the robot as shown in fig13.

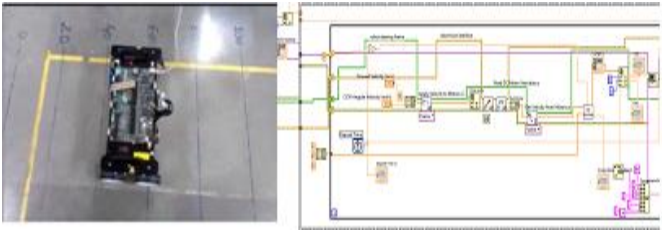

Fig13: Testing of Robot for positional accuracy and its LabVIEW code

The flowchart shown in fig14. describes the flow of system

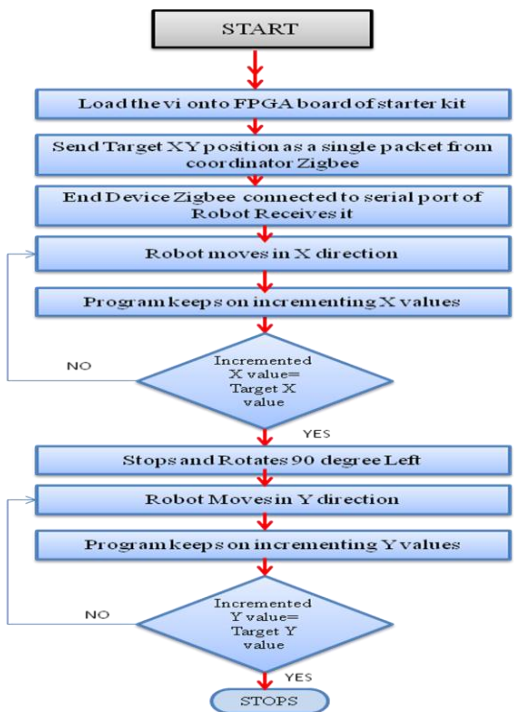

Fig14: Process flow

E. SOFTWARE DEVELOPMENT

A simple graphical user interface is been made in Processing IDE using controlp5 library as shown in fig 15. ControlP5 is a GUI and controller library for processing that can be used in application, applet and android mode. Controllers such as Sliders, Buttons, Toggles, Knobs, Textfields, RadioButtons, Checkboxes amongst others are easily added to a processing sketch. They can be arranged in separate control windows, and can be organized in tabs or groups [11]

3 controls are added named "**Temperature in Lab1**", "**Temperature in Lab2**" and "**Robot position**", such that when user clicks on 1st control, temperature reading of that Lab will be shown, when user clicks on 2nd control, temperature reading of another lab will be shown and so on. The position of Robot can also be viewed dynamically as and when robot keeps on moving. The values are also been logged in Excel file permanently for later usage as shown in fig16.

In this snapshot robot has covered 5m distance from its starting position in x direction and is currently at 3m distance in y direction.

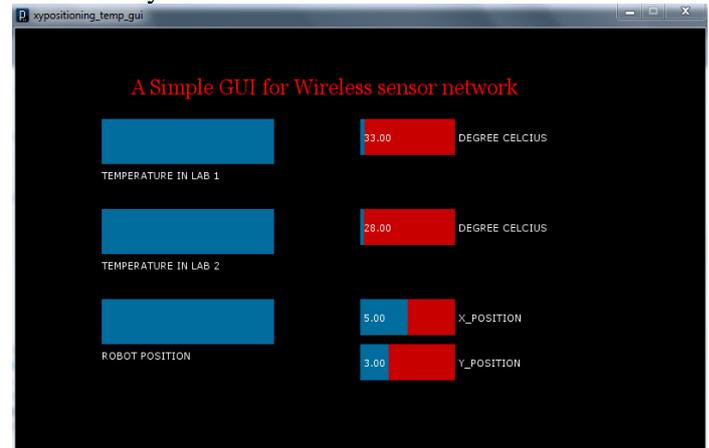

Fig15: GUI

| | A | B | C | D | E | F |
|---|---|---|---|---|---|---|
| 1 | Date and Time | millis sinc | Temp in Lab1 | Temp in Lab2 | Robot's position_X | Robot's position_Y |
| 2 | 1/23/2013/18:8:19 | 2093 | 32 | 28 | 1 | 0 |
| 3 | 1/23/2013/18:8:19 | 2244 | 32 | 28 | 1 | 0 |
| 4 | 1/23/2013/18:8:19 | 2259 | 32 | 28 | 1 | 0 |
| 5 | 1/23/2013/18:8:19 | 2283 | 32 | 22 | 1 | 0 |
| 6 | 1/23/2013/18:8:19 | 2665 | 31 | 23 | 1 | 0 |
| 7 | 1/23/2013/18:8:20 | 3170 | 32 | 22 | 1 | 0 |
| 8 | 1/23/2013/18:8:20 | 3667 | 31 | 22 | 1 | 0 |
| 9 | 1/23/2013/18:8:21 | 4182 | 33 | 22 | 2 | 0 |
| 10 | 1/23/2013/18:8:21 | 4664 | 33 | 23 | 2 | 0 |
| 11 | 1/23/2013/18:8:22 | 5181 | 34 | 24 | 2 | 0 |
| 12 | 1/23/2013/18:8:22 | 5663 | 32 | 26 | 2 | 0 |
| 13 | 1/23/2013/18:8:23 | 6164 | 32 | 28 | 2 | 0 |
| 14 | 1/23/2013/18:8:24 | 7182 | 29 | 28 | 3 | 0 |
| 15 | 1/23/2013/18:8:24 | 7664 | 29 | 29 | 3 | 0 |

Fig16: Real time data storage in Excel file

IV  APPLICATIONS

1) Wireless control and monitor of Robot to reach at target destination.
2) Transportation of object by attaching gripper on its front end as shown in fig 17.

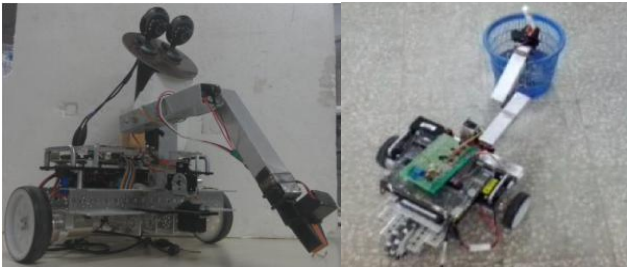
Fig17: pick and place of object

## V SUMMARY AND CONCLUSION

In this paper we developed a prototype of WSN by deploying one temperature node in Electrical Lab and another node in Cryogenic Lab of our institution for temperature monitoring. On the other side of the network a starter kit robot was given its target X-Y position wirelessly and its real time position is been monitored at some remote terminal. Two xbee routers were also used between the labs to extend the range of network and route the packets from xbee end devices to coordinator xbee. The Approximate distance of each node and robot from coordinator node is given in the table below. It also shows the approx delay in receiving of packet from end nodes to the coordinator.

| SENSOR | DISTANCE FROM COORDINATOR | APPROX DELAY |
|---|---|---|
| Node 1 | 17m | 380ms |
| Node 2 | 11m | 312ms |
| Robot | 5m | 170 ms |
| 1 Router | 14m | - |
| 2 Router | 8m | - |

ACKNOWLEDGMENT

This work was carried out in Institute for Plasma Research (IPR), Gandhinagar.